\documentclass[journal]{IEEEtran}

\usepackage[usenames]{color}
\definecolor{bluecolor}{RGB}{0,0,255}
\usepackage{subfigure}

\ifCLASSINFOpdf
  \usepackage[pdftex]{graphicx}
  \usepackage{subfigure}
\else
\fi
\usepackage{algpseudocode}
\usepackage[ruled,vlined]{algorithm2e}
\usepackage{hyperref}

\begin{document}
%
\title{Temporal Heterogeneity Improves \\ Speed and Convergence in Genetic Algorithms}
%
%
%

\author{Yoshio Martinez, 
        Katya~Rodriguez, 
        and~Carlos~Gershenson
\thanks{Authors are with Universidad Nacional Aut\'onoma de M\'exico.}
\thanks{C. Gershenson is also with Lakeside Labs GmbH, Lakeside Park B04, 9020 Klagenfurt am Wörthersee, Austria. email: cgg@unam.mx}
}

%
%

\markboth{Journal of \LaTeX\ Class Files,~Vol.~14, No.~8, December~2020}%
{Shell \MakeLowercase{\textit{et al.}}: Bare Demo of IEEEtran.cls for IEEE Journals}

%



\maketitle

\begin{abstract}
Genetic algorithms have been used in recent decades to solve a broad variety of search problems. These algorithms simulate natural selection to explore a parameter space in search of solutions for a broad variety of problems. In this paper, we explore the effects of introducing temporal heterogeneity in genetic algorithms. In particular, we set the crossover probability to be inversely proportional to the individual's fitness, \emph{i.e.}, better solutions change slower than those with a lower fitness. As case studies, we apply heterogeneity to solve the $N$-Queens and Traveling Salesperson problems. We find that temporal heterogeneity consistently improves search without having prior knowledge of the parameter space.
\end{abstract}


%
\IEEEpeerreviewmaketitle

\section{Introduction}
%
%
%
%

\IEEEPARstart{E}{ngineering} is an area full of challenges which require efficient solutions. Often, due to the complexity of a problem, it is not feasible to find an optimal solution. Thus, it is desirable to find a good enough solution in the shortest possible time. Still, there is a plethora of different search algorithms that can explore a state space to find solutions. As the variety of state spaces to explore is infinite, an open problem lies in finding the parameters or methods that will allow an algorithm to find reliably and consistently a good solution for any given state space.

Genetic algorithms (GA) allow us to find solutions to various problems (combinatorial, regression, optimization) and are inspired by natural selection. These algorithms evolve a population in each generation and individuals are evaluated with a loss function to evaluate the progress of a population of solutions~\cite{10.5555/534133,Mitchell1996}. GA have two main parameters. One is the crossover probability: how likely two solutions are to be combined (mate) to produce new solutions (offspring). The other is the mutation probability: how likely a solution will have random changes.

In this work, we explore the effect of adjusting crossover probabilities heterogeneously. That is, different individuals will have different probabilities (depending on their fitness).
A similar exploration has been made related to mutation probabilities~\cite{Cervantes}. Two combinatorial optimization problems are used for testing the proposed genetic algorithm. The first one is the well-known $N$-queens problem which was proposed by Max Bezel in 1848 and belong to the NP-Complete problems \cite{reinasNP}, the number of solutions to evaluate with N-queens is $N!$, which shows that when $N$ increases the problem becomes intractable when using greddy methods. This aims to place $N$ queens on a $N$ by $N$ chess board, where no queen attacks any other, \emph{i.e.}, there is only one queen in each column, row, and diagonal. The second problem is the Traveling Salesman Problem (TSP), formally defined as follows~\cite{TSP,Sanchez2018}:

The TSP is a combinatorial problem with the objective of finding the path of the shortest length (or the minimum cost) on an undirected graph that represents cities or nodes to be visited. The traveling salesman begins at one node, visits all other nodes consecutively only once, and finally returns to the starting point.  In  other  words,  given  $n$  cities  $\displaystyle{\{c_{1}, c_{2}, ..., c_{n}\}}$, and their permutations $\displaystyle{\{ \sigma_{1}, \sigma_{2}, ..., \sigma_{\frac{(n-1)!}{2}} \}}$, the goal is to find $\displaystyle{\sigma_{i}}$ such that the sum of all Euclidean distances between each node an its successor is minimized.
The successor of the last node in the permutation is the first one, therefore with $n$ nodes or cities, we'll have $(n-1)!/n$ different paths. The Euclidean distance $d$ between any two cities with coordinates $\displaystyle{(x_{1}, y_{1})}$ and $\displaystyle{(x_{2},y_{2})}$ is calculated by:\newline

\begin{equation}
d = \sqrt{(x_{1} - x_{2})^2+(y_{1} - y_{2})^2}
\end{equation}

\hfill

And the path length $l$ to be minimized is calculated as follows:\newline

\begin{equation}
l= \left( \sum_{n=1}^{N-1}d(c_{n}, c_{n+1} )\right)  + { d(c_{N},c_{1}) }
\end{equation}

\hfill

Thus, the aim of this paper is to evaluate the effect of temporal heterogeneity in genetic algorithms for solving these two combinatorial  problems.

\section{Related Work}

There are already several works related to genetic algorithms for solving the $N$-Queens problem. Bozikovic, \emph{et al.}~\cite{AGP} solved the problem  with the help of a parallel genetic algorithm. Farhan, \emph{et al.}~\cite{nQueen} found the 92 solutions for the 8-Queens.
There is also related work to obtain the positions of the queens by chaining solutions for small values of $N$ \cite{ABRAMSON1989649}.
In the same way, there are several works that solve this problem with other methodologies, such as Hu, \emph{et al.}~\cite{pso} who used Particle Swarm Optimization (PSO) to treat this problem. The literature addressing the TSP is vast, \emph{e.g.}~\cite{DORIGO,doi:10.1063/1.5039131,10.1023/A:1006529012972,khanfoziatsp}. However, in this work, the aim is not to find all the possible solutions that a problem has (Table~\ref{tab:solutions} lists the total number of solutions, possible combinations ($N!$), and the probability that a random configuration is a solution for $N$ $<=$ $20$), but to accelerate the search process using a smaller amount of calculations both in the crossover and the evaluation stages of the GA.\newline

\begin{table}[h]
\renewcommand{\arraystretch}{2}
\caption{Number of different solutions for the $N$-queens problem.}
\label{tab:solutions}
\centering
\begin{tabular}{|c|c|c|c|}
\hline

$N$ & $solutions$ & $combinations$ &
\begin{minipage}{1.7cm}
\begin{center}
               $solutions$\\ $/$ \\$combinations$
\end{center}
    
\end{minipage} \\
\hline
1 & 1 & 1 & 1 \\
2 & 0 & 2 & 0 \\
3 & 0 & 6 & 0 \\
4 & 2 & 24 & 0.0833 \\
5 & 10 & 120 & 0.0833 \\
6 & 4 & 720 & $5.5\times10^{-3}$ \\
7 & 40 & 5,040 &  $\approx7.9365\times10^{-3}$ \\
8 & 92 & 40,320 &  $\approx2.2817\times10^{-3}$ \\
9 & 352 & 362,880 &  $\approx9.7001\times10^{-4}$ \\
10 & 724 & 3628,800 &  $\approx1.9951\times10^{-4}$ \\
11 & 2,680 & $\approx399\times10^{5}$ &  $\approx6.7139\times10^{-5}$ \\
12 & 14,200 & $\approx4,790\times10^{5}$ &  $\approx2.9644\times10^{-5}$ \\
13 & 73,712 & $\approx62,270\times10^{5}$ & $\approx1.1837\times10^{-5}$ \\
14 & 365,596 & $\approx871,782\times10^{5}$ & $\approx4.1936\times10^{-6}$ \\
15 & 2,279,184 & $\approx1,307\times10^{9}$ & $\approx1.7429\times10^{-6}$ \\
16 & 14,772,512 & $\approx20,922\times10^{9}$ & $\approx7.0604\times10^{-7}$ \\
17 & 95,815,104 & $\approx355,687\times10^{9}$ & $\approx2.6938\times10^{-7}$ \\
18 & 666,090,624 & $\approx640,237\times10^{10}$ &  $\approx1.0403\times10^{-7}$ \\
19 & 4,968,057,848 & $\approx1.21645\times10^{26}$ &  $\approx4.084\times10^{-8}$ \\
20 & 39,029,188,884 & $\approx2.43290\times10^{27}$ &  $\approx1.6042\times10^{-8}$ \\
\hline
\end{tabular}
\end{table}

Previously, a study entitled ``Rank Diversity of Languages: Generic Behavior in Computational Linguistics''~\cite{ranking} observed that the most important elements in a system change more slowly than the less important ones. This research assigned a rank to a word depending on its frequency, where a rank ``1'' is the lowest, meaning that the word was the most frequently used in a given year, and words with a very low frequency were assigned a high rank. The aforementioned behavior was found: some words that were repeated a lot followed that trend for many years and changed their rank very slowly or not at all, while seldom used words changed their rank considerably every year. In a similar way, in another investigation entitled ``Genetic Temporal Features of Performance Rankings in Sport and Games”~\cite{Morales2016}, the same behavior was observed: successful athletes or teams who remained at a lower rank remained for longer time, while less successful athletes or teams (with higher ranks) changed their rank very quickly. The hypothesis is that such a temporal heterogeneity provides complex systems with a \emph{balance} between stability and variability~\cite{Langton1990,Kauffman1993}, also known as criticality~\cite{christensen2005complexity,Roli2018}. Still, homogeneous models restrict this balance to a phase transition, making it difficult to find parameters and solutions without prior knowledge of a problem, or requiring other mechanisms to guide the evolution of parameters towards criticality~\cite{BTW1987,TorresSosa2012}.\newline

This encourages us to explore the effect of heterogeneity on genetic algorithms, given the evidence that criticality favors evolvability~\cite{packard1988adaptation}. In the methods section we detailed how this approach is included in the genetic algorithms.

\section{Methods}



Our idea is to cross less fit individuals in the population more quickly than fitter individuals. Thus, those with a high fitness value will have a lower probability of mixing with other individuals. As it will be shown, this results a decrease in the number of calculations carried out, allowing the search process to be accelerated.

\subsection{Genetic Algorithms}

The proposed GA has a structure very similar to that used by previous work~\cite{AGP}.
Once an initial population has been generated, a processes for each individual in the population is created. This allows an individual to be independently evaluated. The same applies for the crossover. This is not the case for mutation, since this genetic operator does not require as many instructions as the other two processes mentioned above. 
%
%


\subsection{$N$-queens}

\begin{figure}[h]
\centering
\subfigure[8 Queens]{\includegraphics[scale=0.3]{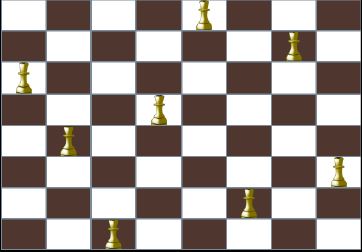}
\label{fig:boards8}}
\subfigure[10 Queens]{\includegraphics[scale=0.3]{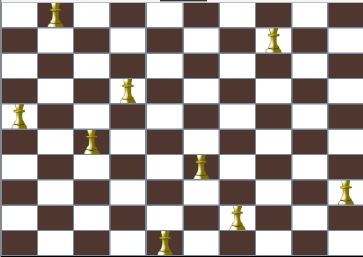}}
\subfigure[20 Queens]{\includegraphics[scale=0.6]{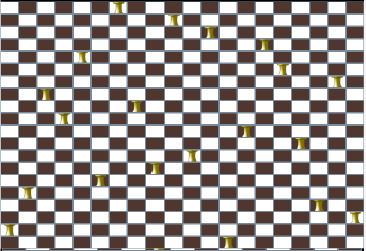}}
\caption{Solutions to 8, 10 and 20 queens.}
\label{fig:boards}
\end{figure}

The goal is to place $N$ queens on a $N$ by $N$ chess board. We will have a population $V$ where each $v_{i}$ represents an individual in the population. To place a queen on the board with the configuration of an individual, the columns of the board are taken as the position $i$ in the array and the rows by the number contained in such position. Below an individual solution to the problem of the 8 queens is given and its placement on the board is shown in Figure~\ref{fig:boards8}:\newline

$v_{i} = \{2,4,7,3,0,6,1,5\}$

\hfill

The initial population of individuals consists of randomly generated vectors without repeated numbers per individual. Here the objective is to minimize the number of intersections between queens across the board (Algorithm~\ref{algo:Nqueens}): zero implies that there are no intersections and thus the problem is solved.\\

\begin{algorithm}[h]
\SetAlgoLined
\KwResult{Return x}
 initialization x\;
 \For{$g=1,2..., v$}{
    \For{$i=1,2..., g$}{
      \If{$v_{i}=v_{g}$ or $abs(v_{g}-v_{i})=abs(i-g)$}{
        $x=x+1$\;
    }
  }
 }
\caption{Objective function of $N$-queens.}
\label{algo:Nqueens}
\end{algorithm}

To calculate the probability of crossing over a couple of individuals $v_{a}$ and $v_{b} $, the following steps are performed:

\hfill

\begin{enumerate}
    \item The entire population is evaluated and fitness is obtained for each individual ${\bf\Phi}(v_{i})$.
    \item The fitness of the worst individual is estimated in the population ${\bf\tau}$, and finally.
    \item To decide if it is possible to crossover ${\bf\Psi}(v_{a}, v_{b})$, the following operation is evaluated:\newline
\end{enumerate}

\begin{equation}
p(v_{a}) = 1 - \{\Phi(v_{a})/\tau\}.
\end{equation}

\hfil

This operation is performed for $v_{a}$ and $v_{b}$, thus obtaining $w_{a}$ and $w_{b}$ (Algorithm~\ref{algo:cross}). This simple operation allows us to slowly crossover the best suited individuals and the worst are constantly crossing over.\\

\hfill

\begin{algorithm}[h]
\SetAlgoLined
\KwResult{Return $V$}
\If{($1 - \{\Phi(v_{a})/\tau\}) <= random$}{
$w_{a} = \Psi(v_{a},v_{b})$
}
\If{$ (1 - \{\Phi(v_{b})/ \tau \}) <= random$}{
 $w_{b} = \Psi(v_{b},v_{a})$
}
\caption{Crossover for $N$-queens and TSP.}
\label{algo:cross}
\end{algorithm}

\subsection{ Traveling Salesperson Problem }

\begin{figure}[h]
    \centering
    \includegraphics[width=3in]{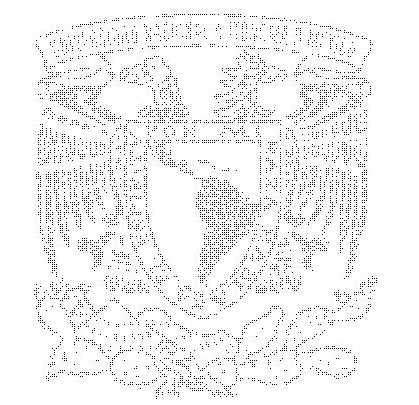}
    \caption{Image with 7,000 points based on the logo of our university.}
    \label{fig:unamLogo}
\end{figure}

The proposed genetic algorithm will be applied to solve a case of the problem of the traveling salesperson, in the same way as for the problem of the $N$-queens. Here, vectors with integers values are generated, where each position of the vector corresponds to a point of Figure~\ref{fig:unamLogo}. \newline

Then, all points have to be visited only once (without repetitions) to find the combination of nodes that give the shortest route. In the same way as with the $N$-queens problem, comparisons between each of the runs considering the distance obtained, considering the execution time and the number of instructions performed.\newline

\section{Experiments}

\subsection{ $N$-queens }
The problem of the $N$-queens is then solved for $N$ values of 20, 40, 60, 80, 100, 150, 200, 250, and 500. Ten or fifty runs were performed for each of the board configurations mentioned above, where the number of solutions per run, runtime, and instructions performed are compared. The configuration used for each board is shown in Table~\ref{tab:nqueens}.\newline

\begin{table}[h]
\renewcommand{\arraystretch}{2.4}
\caption{Configurations used for $N$-queens problem.}
\label{tab:nqueens}
\centering
\begin{tabular}{|c|c|c|c|c|c|}
\hline
$N$ & Population & Crossover & Mutation & Generations & runs \\
\hline
\begin{minipage}{1.5cm}
                20, 40, 60,\\ 80,100, 150
\end{minipage}
& 300 & 0.9 & 0.1 & 500 & 10 \\
\hline
200, 250 & 300 & 0.9 & 0.1 & 1000 & 10 \\
\hline
500 & 500 & 0.9 & 0.1 & 3000 & 50 \\
\hline
\end{tabular}
\end{table}

\subsection{ Traveling Salesperson Problem }
As discussed earlier, the classical genetic algorithm will be tested and compared against a genetic algorithm using temporal heterogeneity. For the case of the TSP, only one run of the genetic algorithm is executed using the parameters shown in Table~\ref{tab:TSP}.

\begin{table}[h]
\caption{Configuration used for the TSP.}
\label{tab:TSP}
\centering
\begin{tabular}{|c|c|c|c|c|c|}
\hline
Points & Population & Crossover & Mutation & Generations\\
\hline
7000 & 300 & 0.9 & 0.1 & 80000 \\
\hline
\end{tabular}
\end{table}

\section{ Results}
In this section, our results obtained are shown, starting with the $N$-queens problem, followed by the TSP. A personal computer with an Intel core i7 and 16 GB of RAM was used.\newline

\subsection{ $N$-queens }
Tables~\ref{tab:solth} and~\ref{tab:solwth} show the results for the heterogeneous and the classical (homogeneous) genetic algorithms, respectively. The first column indicates the number of queens to be placed on the board, the ``Resolved'' column indicates the number of times (of the 10 or 50 runs of the algorithm) that the algorithm reaches a valid dashboard configuration. On average, temporal heterogeneity leads to a higher percentage of runs that find a solution.\newline

The number of average instructions shows those required for crossover, which is the part that we modified. These are drastically reduced with temporal heterogeneity.

Finally, the average time in minutes for runs with $N$ is detailed. Consistently, temporal heterogeneity reduces total computation time.

\begin{table}[h]
\caption{Solutions with temporal heterogeneity.}
\label{tab:solth}
\centering
\begin{tabular}{|c|c|c|c|}
\hline
N & Resolved & Crossover instructions & Time \\
\hline
20 & 10 & 203,106.7 & 0.54 M \\
\hline
40 & 10 & 1,179,512.4 & 2.03 M \\
\hline
50 & 10 & 2,319,122.2 & 2.45 M \\
\hline
60 & 10 & 3,199,086.1 & 3.22 M \\
\hline
80 & 9 & 8,898,568.0 & 6.34 M \\
\hline
100 & 9 & 13,255,157.2 & 3.8 M \\
\hline
150 & 9 & 29,060,304.6 & 9.3 M \\
\hline
200 & 6 & 55,644,604. & 14.19 M \\
\hline
250 & 7 & 83,096,536.7 & 18.33 M \\
\hline
500 & 26 & 380,562,758.3 & 47.35 M \\
\hline
\end{tabular}
\end{table}

\begin{table}[h]
\caption{Solutions without temporal heterogeneity.}
\label{tab:solwth}
\centering
\begin{tabular}{|c|c|c|c|}
\hline
N & Resolved & Crossover instructions & Time \\
\hline
20 & 10 & 1,359,720.7  & 2.06 M \\
\hline
40 & 10 & 5,133,783.2 & 3.27 M\\
\hline
50 & 9 & 11,408,601.2 & 6.25 M\\
\hline
60 & 10 & 9,276,517.8 & 4.01 M \\
\hline
80 & 10 & 19,565,327.2 & 6.36 M \\
\hline
100 & 8 & 29,796,699.0, & 8.12 M \\
\hline
150 & 6 & 71,354,126.8, & 14.25 M \\
\hline
200 & 9 & 97,183,364.6, & 16.38 M \\
\hline
250 & 6 & 159,084,196.8, & 24.54 M \\
\hline
500 & 21 & 580,915,501.2 & 58.31 M \\
\hline
\end{tabular}
\end{table}

\begin{figure}[h]
    \centering
    \includegraphics[scale = 0.6]{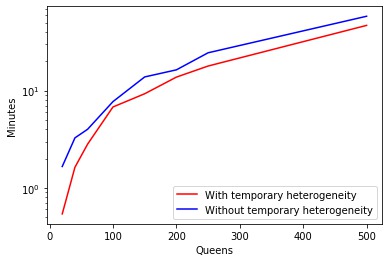}
    \caption{Relation Queens and Time}
    \label{rel:queentime}
\end{figure}

Figure~\ref{rel:queentime} allows to visualize the effect of temporal heterogeneity on computation time as queens are increased. The graph seems to indicate that the advantage of temporal heterogeneity is maintained as the complexity of the problem increases. This suggests that the effect of temporal heterogeneity could still have good results if the number of queens is increased further.\newline

\begin{figure}[h]
    \centering
    \includegraphics[scale = 0.6]{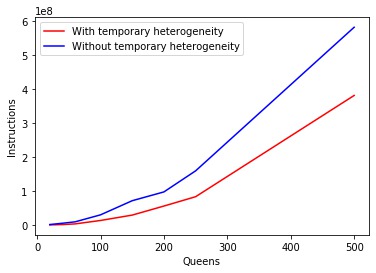}
    \caption{Relation Queens and Instructions}
    \label{rel:queeninstr}
\end{figure}

Having the fittest individuals without crossing over across generations and promoting crossover of the unfit seems to allow local optimal jumping and moving to better regions in the search space. Also, not crossing over the entire population allows to decrease the execution time of the algorithm and the instructions that are carried out, as the evaluation of the objective function is performed less times (Figure~\ref{rel:queeninstr}).

\newpage

These results encourage us to try more queens, thus solving the problem for 1000 queens. Running once for each algorithm for 6000 generations and a population of 300, the following results were obtained (Table~\ref{tab:1000queens}):

\begin{table}[h]
\caption{1000-Queens.}
\label{tab:1000queens}
\centering
\begin{tabular}{|c|c|c|c|}
\hline
Heterogeneity & Resolved & Crossover instructions & Time (minutes) \\
\hline
Yes & 1 & 1,496,961,310  & 191.31 \\
\hline
No & 0 & 2,170,172,477  & 256.34 \\
\hline
\end{tabular}
\end{table}

\hfill

It was interesting to notice that, the genetic algorithm that uses temporal heterogeneity, solves the problem of 1000 queens in just over 3 hours. On the other hand, the classical algorithm did not manage to solve the problem in 6000 generations.\\

\begin{figure}[h]
\centering
\subfigure[1000 Queens, Generation0]{\includegraphics[scale=0.22]{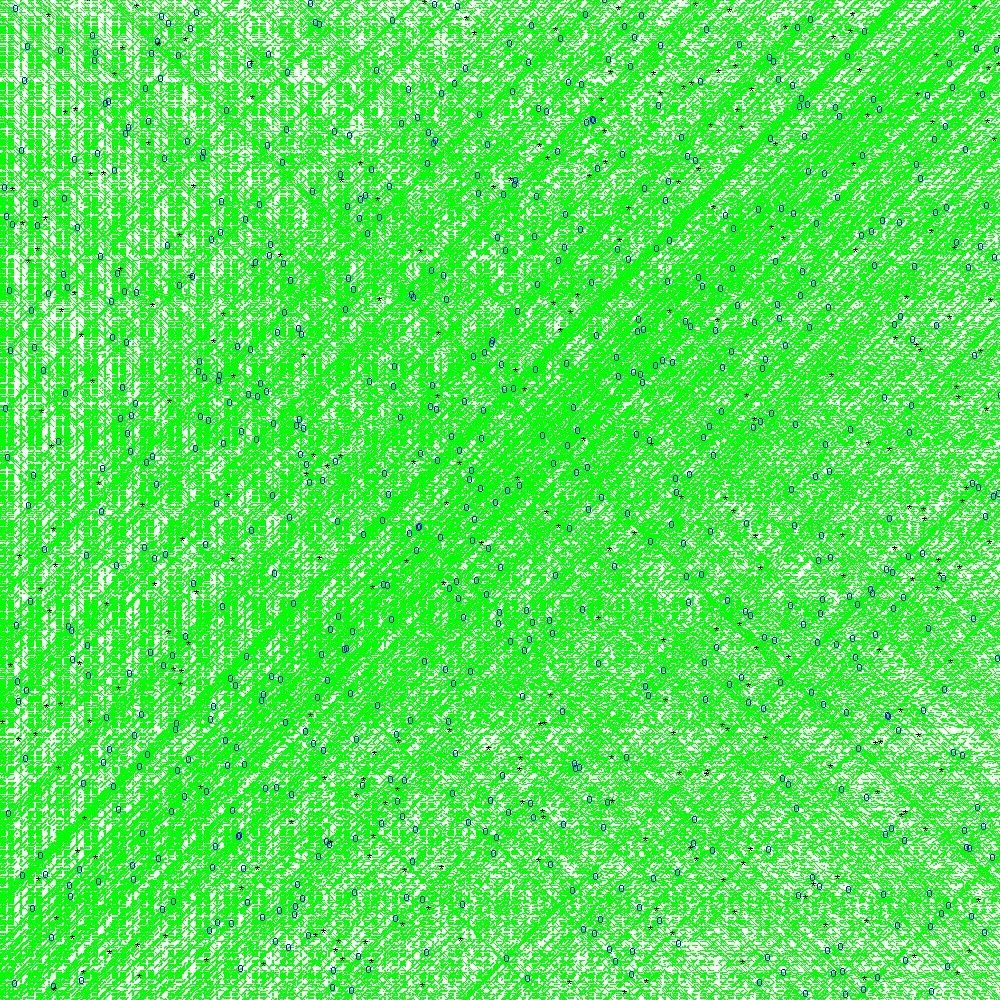}
\label{fig:gen0}}

\subfigure[Solution 1000 Queens with heterogeneity]{\includegraphics[scale=0.22]{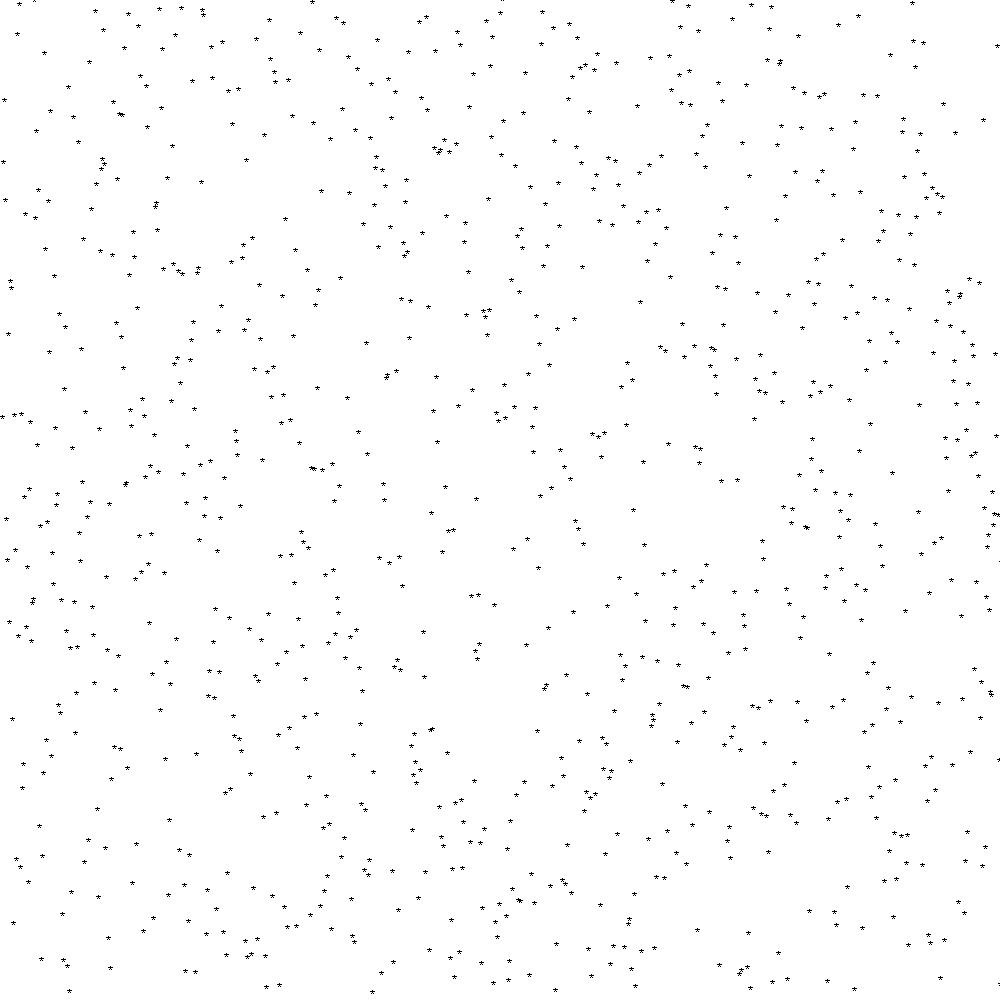}
\label{fig:sol1000}}

\subfigure[Solution 1000 Queens without heterogeneity]{\includegraphics[scale=0.22]{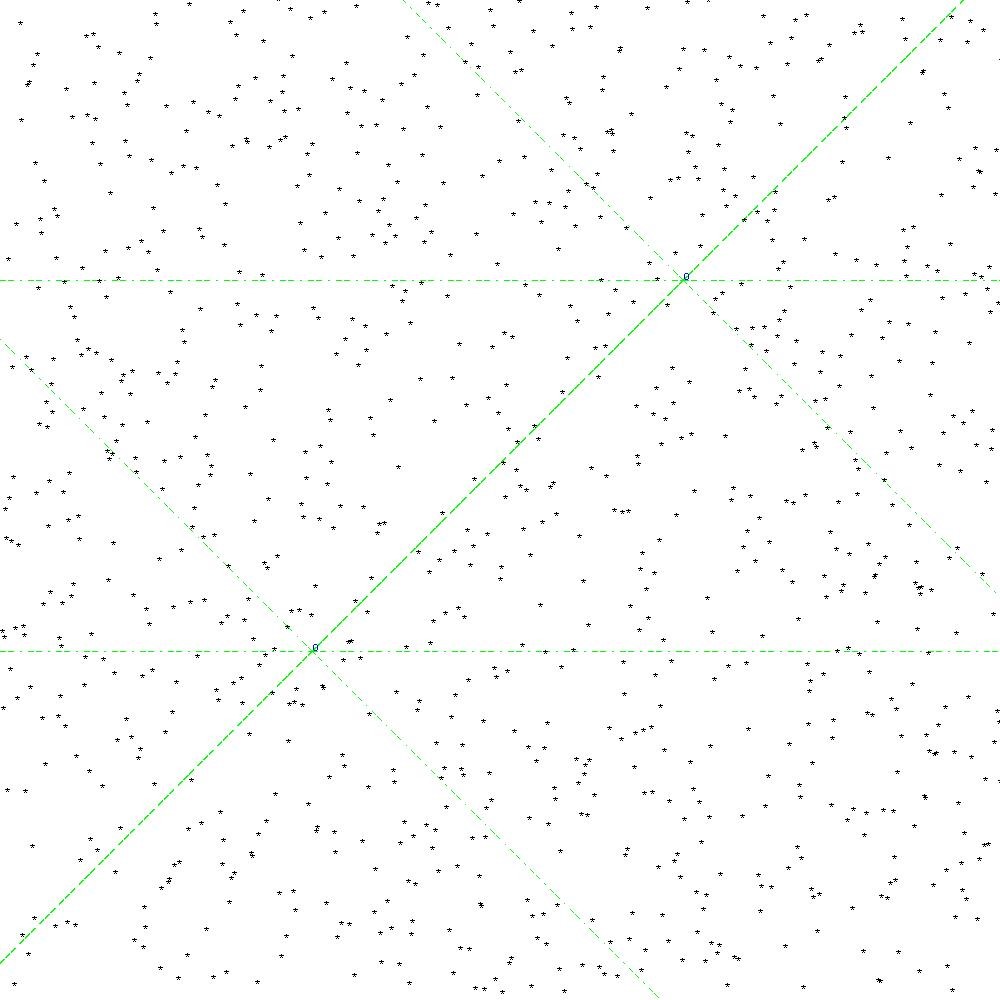}
\label{fig:solerr}}
\caption{Green lines indicate intersection in queens, \textbf{*} indicate a well positioned queen, \textcolor{bluecolor} {\textbf {0}} indicates an attacked queen.}
\end{figure}

In Figure~\ref{fig:gen0}, an individual of the population of 1000 queens in generation 0 is shown, where it is obviously not a correct solution. The queens that are well located are represented with a small asterisk; on the other hand, the queens that are under attack are represented with a blue circle, and green lines are drawn from it in all directions except vertical. In Figure~\ref{fig:sol1000}, the solution to 1000 queens using the heterogeneous algorithm is presented. On the other hand, in Figure~\ref{fig:solerr}, the best individual that the classic algorithm generated is given, where it is clearly seen that it is not a solution to the problem due to having intersections (attacks).

\subsection{Traveling Salesperson Problem}
The results obtained using the genetic algorithm with and without heterogeneity applied to the TSP with 7,000 nodes are shown as follows.

\begin{table}[h]
\caption{TSP Performance}
\label{tab:resultsp}
\centering
\begin{tabular}{|c|c|c|c|}
\hline
Heterogeneity & Distance & Crossover instructions & Time (hours) \\
\hline
Yes & 24,032.29 & 211,337.7   & 87.24 \\
\hline
No & 35,395.33 & 192,617.8  & 94.89 \\
\hline
\end{tabular}
\end{table}

\hfill

\begin{figure}[h]
    \centering
    \includegraphics[scale = 0.5]{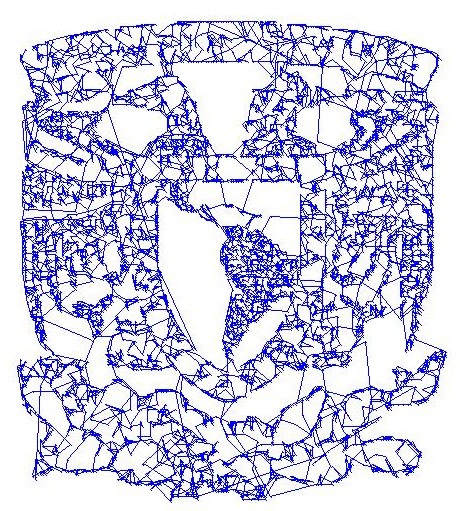}
    \caption{Solution TSP with heterogeneity}
    \label{fig:solTSP}
\end{figure}

Table~\ref{tab:resultsp} shows the distance, number of instructions and time.
Figure~\ref{fig:solTSP} is the result obtained by running the genetic algorithm using temporal heterogeneity for 80,000 generations, that although the algorithm made more instructions in the crossover operation, it could obtain a considerably better result in slightly less time, suggesting that temporal heterogeneity could be valuable to solve search problems more complex than those studied here. Still, this goes beyond the scope of this paper will be the focus of future work.

\newpage

\newpage

\section{Conclusion}
As it was shown in the problems studied here, temporal heterogeneity in genetic algorithms leads to better and faster results due to the fact that it performs less operations with the individuals, which allows to accelerate the search process. The rationale behind our method is similar to the simulated annealing algorithm \cite{recocidosimulado}. Both algorithms advance with smaller steps when they are closer to the solution. Still, in simulated annealing, it is not trivial to find the optimal cooling constant, which varies with the particular problem. On the contrary, with the proposed method, we do not need any additional parameter to achieve good results in different instances of the combinatorial problems studied here.\newline

Nevertheless, the benefits of temporal heterogeneity in genetic algorithms do not mean that it will be universally beneficial for all possible problems \cite{nolunch}. Thus, it is necessary to test temporal heterogeneity in different problems to truly measure its usefulness. In this paper we showed that temporal heterogeneity helps problems with integer coding, such as $N$-queens and TSP. As a future work, we will evaluate the potential benefits of temporal heterogeneity in real coding problems and stochastic environments. We have already trained a deep neural network to learn how to play Atari games by reinforcement with the help of temporal heterogeneity, obtaining encouraging preliminary results.\newline

It would also be interesting to test whether temporal heterogeneity in mutations would provide additional benefits to that found in crossover. Also, the precise heterogeneity we used was arbitrary. It would be useful to explore systematically different degrees of heterogeneity, in the search of an ``optimal'' heterogeneity for a particular problem.




%



\ifCLASSOPTIONcaptionsoff
  \newpage
\fi



%

\newpage

\bibliographystyle{plain}
\bibliography{references.bib}

\begin{thebibliography}{10}

\bibitem{ABRAMSON1989649}
Bruce Abramson and Moti Yung.
\newblock Divide and conquer under global constraints: A solution to the
  n-queens problem.
\newblock {\em Journal of Parallel and Distributed Computing}, 6(3):649 -- 662,
  1989.

\bibitem{BTW1987}
Per Bak, Chao Tang, and Kurt Wiesenfeld.
\newblock Self-organized criticality: An explanation of the 1/f noise.
\newblock {\em Phys. Rev. Lett.}, 59(4):381--384, July 1987.

\bibitem{AGP}
M.~Bozikovic, M.~Golub, and Leo Budin.
\newblock Solving n-queen problem using global parallel genetic algorithm.
\newblock In {\em The IEEE Region 8 EUROCON 2003. Computer as a Tool}, pages
  104 -- 107 vol.2, 10 2003.

\bibitem{Cervantes}
Jorge Cervantes-Ojeda and Christopher Stephens.
\newblock A rank-proportional generic genetic algorithm.
\newblock In Theodore Simos, editor, {\em Recent Progress in Computational
  Sciences and Engineering}, pages 71--74. Taylor \& Francis, 05 2006.

\bibitem{christensen2005complexity}
Kim Christensen and Nicholas~R Moloney.
\newblock {\em Complexity and criticality}.
\newblock World Scientific Publishing Company, 2005.

\bibitem{ranking}
Germinal Cocho, Jorge Flores, Carlos Gershenson, Carlos Pineda, and Sergio
  S{\'a}nchez.
\newblock Rank diversity of languages: Generic behavior in computational
  linguistics.
\newblock {\em PLOS ONE}, 10(4):1--12, 04 2015.

\bibitem{reinasNP}
Kelly Crawford.
\newblock Solving the n-queens problem using genetic algorithms.
\newblock 2:1039--1047, 01 2010.

\bibitem{DORIGO}
Marco Dorigo and Luca~Maria Gambardella.
\newblock Ant colonies for the travelling salesman problem.
\newblock {\em Biosystems}, 43(2):73 -- 81, 1997.

\bibitem{nQueen}
Ahmed~S. Farhan, Wadhah~Z. Tareq, and Fouad~H. Awad.
\newblock Solving n queen problem using genetic algorithm.
\newblock {\em International Journal of Computer Applications}, 122(12):11--14,
  July 2015.

\bibitem{doi:10.1063/1.5039131}
Chunhua Fu, Lijun Zhang, Xiaojing Wang, and Liying Qiao.
\newblock Solving tsp problem with improved genetic algorithm.
\newblock {\em AIP Conference Proceedings}, 1967(1):040057, 2018.

\bibitem{10.5555/534133}
David~E. Goldberg.
\newblock {\em Genetic Algorithms in Search, Optimization and Machine
  Learning}.
\newblock Addison-Wesley Longman Publishing Co., Inc., USA, 1st edition, 1989.

\bibitem{pso}
Xiaohui Hu, Russell Eberhart, Yuhui Shi, and Yuhui Com.
\newblock Swarm intelligence for permutation optimization: Case study of
  n-queens problem.
\newblock {\em Proceedings of the IEEE Swarm Intelligence Symposium}, 2003, 07
  2003.

\bibitem{Kauffman1993}
S.~A. Kauffman.
\newblock {\em The Origins of Order}.
\newblock Oxford University Press, Oxford, UK, 1993.

\bibitem{khanfoziatsp}
Fozia Khan, Nasiruddin Khan, Dr.~Syed Inayatullah, and Shaikh Nizami.
\newblock Solving tsp problem by using genetic algorithm.
\newblock {\em International Journal of Basic \& Applied Sciences}, 9, 08 2010.

\bibitem{recocidosimulado}
S.~Kirkpatrick, C.~D. Gelatt, and M.~P. Vecchi.
\newblock Optimization by simulated annealing.
\newblock {\em Science}, 220(4598):671--680, 1983.

\bibitem{Langton1990}
Christpher~G. Langton.
\newblock Computation at the edge of chaos: Phase transitions and emergent
  computation.
\newblock {\em Physica D}, 42:12--37, 1990.

\bibitem{10.1023/A:1006529012972}
Pedro Larranaga, Cindy Kuijpers, R.~Murga, I.~Inza, and S.~Dizdarevic.
\newblock Genetic algorithms for the travelling salesman problem: A review of
  representations and operators.
\newblock {\em Artificial Intelligence Review}, 13:129--170, 01 1999.

\bibitem{Mitchell1996}
Melanie Mitchell.
\newblock {\em An Introduction to Genetic Algorithms}.
\newblock MIT Press, 1996.

\bibitem{Morales2016}
Jos{\'e}~A. Morales, Sergio S{\'a}nchez, Jorge Flores, Carlos Pineda, Carlos
  Gershenson, Germinal Cocho, Jer{\'o}nimo Zizumbo, Rosal{\'\i}o~F.
  Rodr{\'\i}guez, and Gerardo I{\~{n}}iguez.
\newblock Generic temporal features of performance rankings in sports and
  games.
\newblock {\em EPJ Data Science}, 5(1):33, 2016.

\bibitem{packard1988adaptation}
Norman~H Packard.
\newblock Adaptation toward the edge of chaos.
\newblock In J.~A.~S. Kelso, A.~J. Mandell, and M.~F. Shlesinger, editors, {\em
  Dynamic Patterns in Complex Systems}, pages 293--301. World Scientific,
  Singapore, 1988.

\bibitem{Roli2018}
Andrea Roli, Marco Villani, Alessandro Filisetti, and Roberto Serra.
\newblock Dynamical criticality: Overview and open questions.
\newblock {\em Journal of Systems Science and Complexity}, 31(3):647--663,
  2018.

\bibitem{Sanchez2018}
Sergio S\'anchez, Germinal Cocho, Jorge Flores, Carlos Gershenson, Gerardo
  I{\~{n}}iguez, and Carlos Pineda.
\newblock Trajectory stability in the traveling salesman problem.
\newblock {\em Complexity}, 2018:2826082, 2018.

\bibitem{TSP}
Susmita Sur~Kolay, Satyajit Banerjee, and C.~Murthy.
\newblock Flavours of traveling salesman problem in vlsi design.
\newblock In {\em IICAI}, pages 656--667, 01 2003.

\bibitem{TorresSosa2012}
Christian Torres-Sosa, Sui Huang, and Maximino Aldana.
\newblock Criticality is an emergent property of genetic networks that exhibit
  evolvability.
\newblock {\em PLoS Comput Biol}, 8(9):e1002669, 09 2012.

\bibitem{nolunch}
D.~H. {Wolpert} and W.~G. {Macready}.
\newblock No free lunch theorems for optimization.
\newblock {\em IEEE Transactions on Evolutionary Computation}, 1(1):67--82,
  April 1997.

\end{thebibliography}


%

\newpage
\begin{IEEEbiography}[{\includegraphics[width = 1in, height=1.25in,clip]{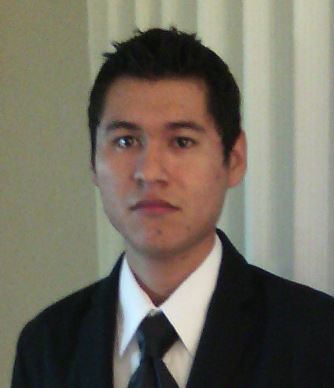}}]{Yoshio Ismael Mart\'inez Ar\'evalo}
Yoshio is a master's candidate at Universidad Nacional Aut\'onoma de M\'exico (UNAM). He has a bachelor's degree in Computer Systems and an interest in Complex Systems, Deep Learning, and Evolutionary Computing. He has extensive experience as software developer in high and low level languages, he is also a fan of applied robotics, blockchain applications, and software acceleration by GPU's.
\end{IEEEbiography}

\begin{IEEEbiography}[{\includegraphics[width = 1in, height=1.25in,clip]{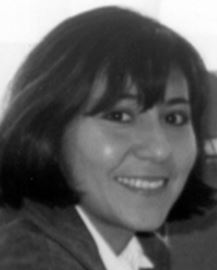}}]{Katya Rodr\'iquez V\'azquez}
K. Rodr\'iguez-V\'azquez received the BEng degree in Computing Engineering from the
National Autonomos University of Mexico (UNAM), in 1994 and the Ph.D from the
University of Sheffield, Sheffield, UK, in 1999.She is currently a full time Research at
the Institute of Applied Mathematics and Systems (UNAM). She has published a
number of papers in international journals and conferences. Her research interests are
in evolutionary computation, multiobjective optimization and its applications.,Dr.
Rodr\'iguez-V\'azquez has been a Member of the Technical Program Committee for
conferences related to evolutionary computation and reviewer of related journals.
\end{IEEEbiography}


\begin{IEEEbiography}[{\includegraphics[width = 1in, height=1.25in,clip]{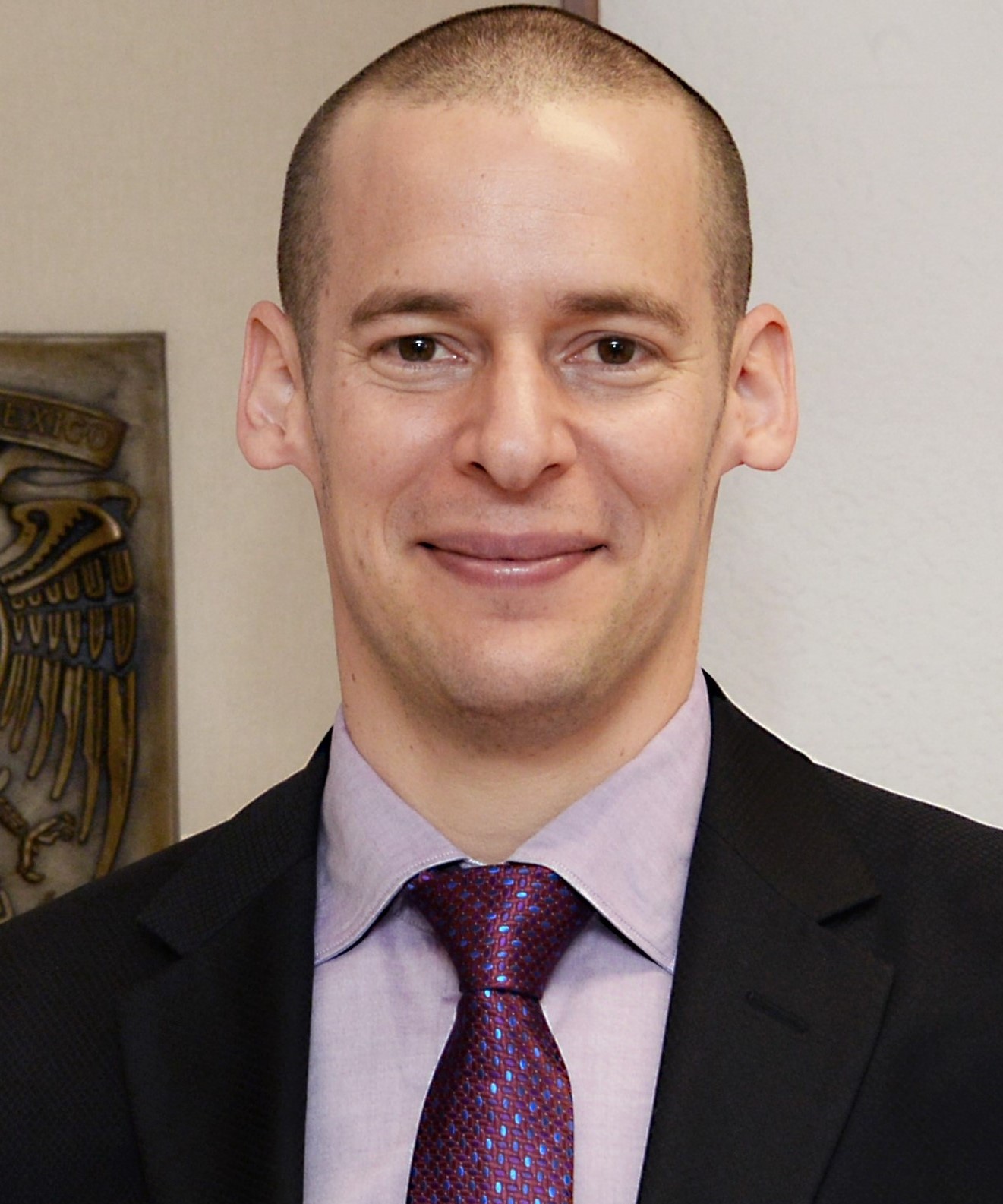}}]{Carlos Gershenson Garc\'ia}
Carlos Gershenson is a tenured, full time research professor at the computer science department of the Instituto de Investigaciones en Matem\'aticas Aplicadas y en Sistemas at the Universidad Nacional Aut\'onoma de M\'exico (UNAM), where he leads the Self-organizing Systems Lab. He is also an affiliated researcher at the Center for Complexity Sciences at UNAM. He is a honorary member at Lakeside Labs, Austria. He was a Visiting Professor at MIT and at Northeastern University (2015-2016) and at ITMO University (2015-2019). He was a postdoctoral fellow at the New England Complex Systems Institute (2007-2008). He holds a PhD  summa cum laude from the Vrije Universiteit Brussel, Belgium (2002-2007). His thesis was on “Design and Control of Self-organizing Systems”. He holds an MSc degree in Evolutionary and Adaptive Systems, from the University of Sussex (2001-2002), and a BEng degree in Computer Engineering from the Fundaci\'on Arturo Rosenblueth, M\'exico. (1996-2001).

	He is Editor-in-Chief of \emph{Complexity Digest}, Associate Editor for the journals \emph{Complexity} and \emph{Frontiers in Robotics and AI}, and member of the Board of Advisors for \emph{Scientific American}. He has received numerous awards, including a Google Research Award in Latin America and the Audi Urban Future Award.
\end{IEEEbiography}




\end{document}